\documentclass[runningheads]{llncs}

\usepackage{bigfoot}
\usepackage{graphicx}
\usepackage{lipsum}
\usepackage{amsmath}
\usepackage[backref=page]{hyperref}
\usepackage{xcolor}

\usepackage[export]{adjustbox}[2011/08/13] 
\newcommand\figscale{1.3}
\usepackage{bbm}

\begin{document}
%
\title{T-DominO}
\subtitle{Exploring Multiple Criteria with Quality-Diversity and the Tournament Dominance Objective}
%
%
\author{Adam Gaier\inst{1} \and
James Stoddart\inst{1} \and
Lorenzo Villaggi\inst{1}\and
Peter J Bentley\inst{1,2}}
\authorrunning{A. Gaier et al.}
%
\institute{Autodesk Research
\and
University College London, London, UK\\
\email{adam.gaier@autodesk.com}}

%
\maketitle              

\begin{abstract}
Real-world design problems are a messy combination of constraints, objectives, and features. 
Exploring these problem spaces can be defined as a Multi-Criteria Exploration (MCX) problem, whose goals are to produce a set of diverse solutions with high performance across many objectives, while avoiding low performance across any objectives. 
Quality-Diversity algorithms produce the needed design variation, but typically consider only a single objective.
We present a new ranking, T-DominO, specifically designed to handle multiple objectives in MCX problems.
T-DominO ranks individuals relative to other solutions in the archive, favoring individuals with balanced performance over those which excel at a few objectives at the cost of the others. 
Keeping only a single balanced solution in each MAP-Elites bin maintains the visual accessibility of the archive -- a strong asset for design exploration.
We illustrate our approach on a set of easily understood benchmarks, and showcase its potential in a many-objective real-world architecture case study.
\end{abstract}
%
 
\begin{figure}[ht]
  \centering
  \vspace{-0.3cm}
  \includegraphics[width=\figscale\textwidth,center]{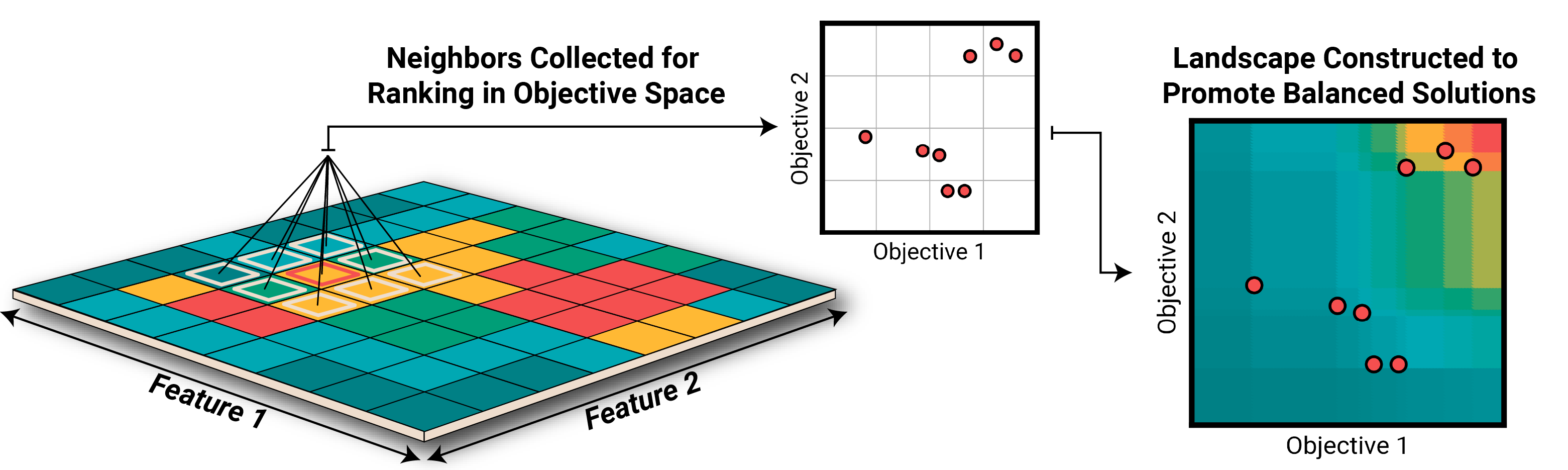}
  \caption
  { 
    \textbf{Calculating the Tournament Dominance Objective (T-DominO)}
  }
  \label{fig:overview}
\end{figure}
\section{Introduction}

Architecture projects must balance a dizzying array of objectives: daylight, views, noise, wind, cost, open spaces, carbon footprint, and ease of construction, to name a few -- along with less easily optimized subjective considerations like aesthetics and comfort. In generative design (GD), where algorithms aid design exploration by producing candidate designs, the desired result is not a single solution, but a variety of high performing options~\cite{bradner2014parameters}. A variety of options is required  because the problem has more than one objective, which means there may be many possible solutions. Perhaps more importantly, a varied choice highlights design concepts to stakeholders and decision makers who then select and modify them according to messy human compromises.
 
Though it may resemble multi-objective optimization (MOO)~\cite{moo}, the problem this work focuses on for GD and similar domains is different.
We define the problem as a Multi-Criteria Exploration (MCX) problem, whose goals are to:

\begin{enumerate}
  \item Produce a catalog of diverse solutions
  \item with high performance across many objectives
  \item while avoiding low performance across any objectives
\end{enumerate}

MCX can be considered an exploratory form of MOO, just as Quality-Diversity (QD)~\cite{cully2017quality,pugh2016quality}, is an exploratory form of single-objective optimization. In contrast to MOO, in MCX we do not strive for either uniform coverage of the Pareto front, nor precise proximity to it. Uniform coverage of the front implies coverage of the extremes of the objective space -- where solutions earn their place in the Pareto front by dominating on only a subset of objectives. These solutions are uninteresting for MCX, as solutions which disregard user preferences by ignoring some objectives are not useful in practice. Proximity to the front is also less important for exploration -- the goal is to generate starting points, not end points. Generated solutions are rarely used without modification, reducing the effort of finding the precise front to an expensive distraction.

The QD approach seems, at first, ideal for solving MCX problems. QD algorithms provide a way of explicitly searching for diversity as defined at a high level by users. Whereas MOO strives for a maximum spread in the objective space, QD searches for spread in a user-defined `feature'\footnote{also referred to in the QD literature as a behavior, descriptor, outcome, or measure} space. As opposed to objectives, these features correspond to different \emph{ways} of solving the problem, not quantities to be minimized. In architecture the number of buildings in a building complex or the distance between them can be explicitly explored with QD in a way that is not possible with MOO.

But QD is designed to explore several features, not to optimize multiple objectives. MAP-Elites~\cite{cully2015robots,mapelites}, the most widely used QD algorithm, divides the feature space and searches for the best solution in each partition. The result is a set of optimized designs organized by high level features; the performance of this collection can then be viewed as heat map projected on to the feature space, illuminating the relationship between features and performance. 

Recent work has proposed combining MOO and QD by computing a Pareto front in every partition~\cite{moo_qd}. In the MCX case this `all of the above' approach is not satisfying -- MAP-Elites' intuitive way of organizing, summarizing, and presenting the results depends on finding a single best solution for each partition. An alternate method of reconciling the approaches must be found.

In this work we leverage the insight that the diversity in objective space produced by MOO mechanisms such as crowding distance~\cite{nsga2}, or reference vectors~\cite{nsga3} are unneeded when diversity is enforced by QD. In QD, users can choose the type of variety to explore, and trade-offs in objectives will naturally arise from those choices. When exploration of the objective space is no longer prioritized, non-domination -- which favors the extremes of the front to the same degree as the center -- ceases to be the most desirable attribute of a solution. In MCX, balanced solutions which perform well on all objectives are preferred. Our second insight is that the `balance' of a solution can be defined in relation to a population, and that the solutions contained in the MAP-Elites archive can act as that population.

Our approach extends MAP-Elites to the exploration of problems with multiple objectives by introducing the Tournament Dominance Objective (T-DominO), which ranks individuals in a population according to an approximation of their distance to the center of the Pareto front. T-DominO awards poor scores to non-dominated solutions at the extremes of the objective space -- those which excel at one objective while doing very poorly at the other -- while those at the center of the front receive the highest scores (Figure \ref{fig:overview}).

Optimizing MAP-Elites according to T-Domino provides an elegant approach to tackle MCX problems.
A simple alternate ranking causes minimal disruption to the core algorithmic machinery while allowing MAP-Elites to discover varied design concepts which balance multiple objectives. Approaches which assume a single objective and single solution in each bin, such as CMA-ME~\cite{cmame}, can still be used. Crucially, T-DominO tackles multiple objectives without sacrificing MAP-Elites' intuitive visualization and analysis of solutions, features, and their interaction with objectives -- the true goal of the algorithm when used for design.


\section{Background}
	\subsection{Generative Design}\label{sec:sec31}	

 In design and architecture, experiments with human-machine collaboration are common~\cite{holzer2007parametric,turrin2011design,gerber2012design,arieff2013new,bradner2014parameters,nagy2017project}. Recent work has demonstrated the viability of GD in real-world applications, from office retrofits~\cite{nagy2017project}, large scale trade-shows~\cite{nagy2017beyond}, to neighborhood scale planning~\cite{nagy2018generative}. The number of conflicting constraints, preferences, and objectives in these projects makes `solving' them an ill-defined and impossible task. Optimization tools are typically used at the \textit{beginning} of the design process rather than the end. Optimization algorithms are not used solve problems, but to \textit{explore} them~\cite{bradner2014parameters,matejka2018dream}. 

The purpose of GD is less optimization and more communication. Search algorithms are used to understand the possibilities and potential of a problem space. Objectives serve as proxies for preferences, goals, and features of interest that are often difficult or impossible to define mathematically. These objectives signify criteria of a good design or ways of counter-balancing those criteria to prevent extreme solutions which are not aligned with designers intent. 

Results are then filtered and categorized in an effort to find qualitatively different design concepts. Typically designs are judged visually first, and only once a set of interesting and varied designs identified is their performance examined. This can be a clumsy process, and for GD to have real success accessibility must be a consideration at every step of the process, including optimization. An intuitive GD approach would not only find a set of solutions which balance performance over several objectives, but explicitly search for the high level diversity that sets solutions apart from each other. This is the goal of MCX.
	\subsection{Exploration and Optimization with Non-Objective Criteria}\label{sec:sec32}	

MOO approaches strive to produce a set of non-dominated solutions that is diverse in the objective space, and as near to the Pareto optimal front as possible~\cite{panichella2019adaptive}, but
the diversity that interests designers is often not in objective space. Other qualities can be induced with `helper' objectives in a process known as multiobjectivization~\cite{knowles2001reducing,handl2008multiobjectivization,jensen2004helper}. Helper objectives can optimize quantities unrelated to performance, such as the type of cross sections in a structural frame~\cite{greiner2007improving}, or the similarity to previous solutions~\cite{mouret2011novelty}, but are still performing minimization. Maximizing or minimizing the number of buildings on a site makes little sense, but understanding the effect of the number of buildings is a valuable insight.

QD approaches such as MAP-Elites~\cite{cully2015robots,mapelites} search for solutions along a continuum of user-defined features, making them ideal for exploration. MAP-Elites has been used for design exploration in domains such as aerodynamics~\cite{sail_aiaa,sail_gecco,sail_ecj,produqd,sphen}, and game design~\cite{alvarez2019empowering,charity2020mech,gonzalez2020finding,gravina2019procedural}, but has been restricted to consideration of a single objective.
MAP-Elites operates by first discretizing the feature space into bins, collectively known as a map or archive. Each bin contains a single solution and its corresponding fitness value. New solutions are created by selecting and varying solutions from the map. These new solutions are evaluated and two values produced: a performance measure and a set of coordinates in the feature space. These coordinates indicate the bin to which the solution belongs. The solution is placed in the bin if it is empty, or if the candidate solution has higher performance than the current occupant of the bin, it replaces it.

The elitist nature of MAP-Elites, with only one solution per bin, puts it at odds with the idea of the Pareto front. A concurrent work~\cite{moo_qd} bridges this gap by introducing a Pareto front in each bin, and replacing fitness tournaments with non-domination. Though this technique is able to find a large set of Pareto fronts, it sacrifices the elegant method of communicating the results. Rather than viewing individual designs and correlations between features and objectives, we are left with a mass of summary statistics -- useful for MOO, but not for MCX. In our work we maintain the the elitist nature of MAP-Elites, and instead replace the Pareto front with an alternate formulation of multi-objective performance.

\section{Method}
\begin{figure}[t]
  \centering
  \vspace{-0.2cm}
  \includegraphics[width=\figscale\textwidth,center]{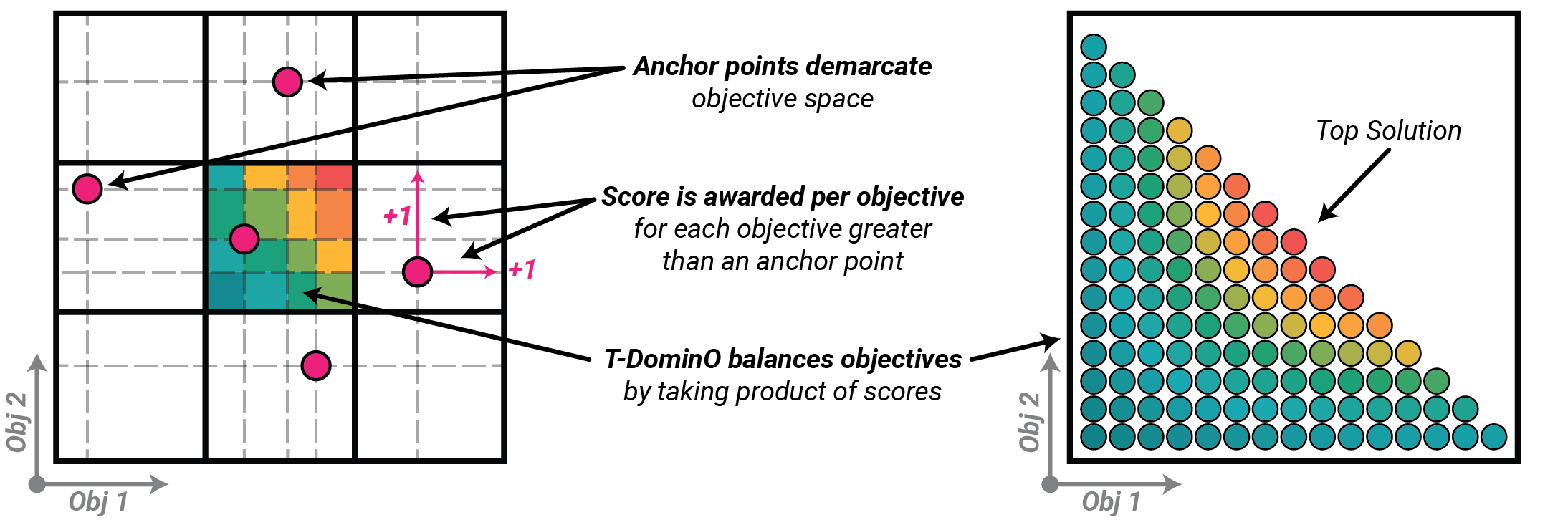}
  \caption
  { 
    \textbf{Using anchor points to calculate T-DominO.} 
  }
  \label{fig:anchor}
\end{figure}

When tackling MCX problems, our interest lies in finding solutions which perform well on all objectives in each region of a QD feature space. The Tournament Dominance Objective (T-DominO), introduced here, ranks solutions according to an approximation of their distance to the center of the front, with the most balanced solutions ranking highest. This approximation is calculated through a series of tournaments between a solution and a set of existing points in objective space, or \textit{anchor points} (Figure \ref{fig:anchor}). An individual is compared to each anchor point on a single objective, and for every anchor point with a lesser or equal objective value one point is awarded. This count is made for every objective, and these counts multiplied. 

The T-DominO score of solution with objective values $x$, compared to a set of anchor points with objective values $A$ is more precisely defined as:
\begin{equation}
    \operatorname{T-DominO}(x, A)=\prod_{n=1}^{objs}\sum_{m=1}^{anchors} f(x_n, A_{mn}),    
        f(x,a)= 
\begin{cases}
    1, & x \geq a\\
    0, & \text{else}
\end{cases}
\end{equation}
where $objs$ and $anchors$ is the number of objectives and anchor points.\footnote{Or in python: \verb|numpy.prod(numpy.sum(objs >= anchors,axis=0))|}

The integration of T-DominO into MAP-Elites can be summarized as follows: 
\begin{enumerate}
  \item a new individual, based on its feature coordinates, is assigned a bin.
  \item a set of anchor points are collected from the history of elites in that bin, the $k$ neighboring bins, and the new individual itself.
  \item the T-DominO of the current elite and the challenger are computed based on these anchor points.
  \item if the challenger has a greater T-DominO score it replaces the current elite, and the objective values of the replaced elite are stored in the bin to serve as a future anchor point.
\end{enumerate}

When individuals in a population are ranked according to T-DominO the result is a ranking from the center of the front outwards (Figure \ref{fig:anchor}, right). This ranking allows the combination of multiple objectives into a single score which rewards solutions with the highest balanced performance, without the need for penalty functions or arbitrary weighting of objectives. 

T-DominO is based around comparisons to anchor points, and the MAP-Elites archive provides a ready source. Existing elites in the archive can be used as a sampling of the objective space, and act as anchor points distributed across both objective and feature space. Selection pressure toward improved T-DominO scores creates high performing solutions which in turn act as anchor points, in a virtuous cycle that leads to ever higher performance.

However, when selection pressure is organized around \textit{only} the current population, cycling can occur. In some circumstances a challenger solution which is better on one objective can replace the current elite, which can in turn be replaced by the original. To prevent this behavior and ensure progress towards better solutions we track the objective values of the previous elites. While each bin continues to contain only a single elite, the objective values of previous elites are maintained to act as anchor points, preventing cycling and creating further refinement of the T-DominO landscape. A simple FIFO buffer of the objective values of a handful of past elites is sufficient.

Neighboring partitions typically have similar performance potential, and so are necessary for creating more fine-grained landscapes, but bins with solutions that dominate all or none of the solutions in the a bin provide no signal to inform selection pressure, and so we can safely limit the anchor points to those contained in the $k$ nearest neighboring bins.

T-Domino allows the simultaneous optimization of several objectives with a single measure, relying on QD diversity mechanisms to prevent convergence on a single point. The output of T-Domino MAP-Elites is ideal for MCX -- an archive with a single best balanced solution in each bin. Having a single solution in each bin allows the effects of solution features on that balance to be easily understood and visualized and so contribute to the understanding of the underlying problem, such as the correlation of features and objectives. The creation of a library of designs organized by high level features of the users choosing provides an ideal set of starting points for further refinement.

\section{Benchmarks}
    \subsection{Setup}

We validate the expected behavior of MAP-Elites with T-DominO on a series of established multi-objective benchmark problems. The purpose of these tests is to validate our claims that T-Domino will:
\begin{enumerate}
    \item Discover high performing, if not optimal, solutions
    \item Produce balanced solutions whose performance does not come at the cost of large trade-offs in a subset of objectives
\end{enumerate}

\subsubsection{Benchmark Functions}\hfill\\
\indent\textit{RastriginMOO}. To judge the performance of T-DominO on Multi-Objective QD problems, we test on a version of RastriginMOO as introduced in~\cite{moo_qd}. The Rastrigin function is a classic optimization benchmark, often used to test QD algorithms because it contains many local minima~\cite{cmame,cully2021multi}. Here it is converted into a multiobjective benchmark by optimizing a pair of Rastrigin functions with shifted centers. We use a 10-D version with constants added so that every discovered bin has a positive effect on the aggregate QD Score. These objectives can be explicitly defined as:
\begin{align}
    \begin{cases}
      f_1(\mathbf{x}) = 200 - (\sum\limits_{i=1}^n [(x_i - \textcolor{blue}{\lambda_1})^2 - 10\cos (2\pi (x_i - \textcolor{blue}{\lambda_1}))]) \\
      f_2(\mathbf{x}) = 200 - (\sum\limits_{i=1}^n [(x_i - \textcolor{blue}{\lambda_2})^2 - 10\cos (2\pi (x_i - \textcolor{blue}{\lambda_2}))])
    \end{cases}       
\end{align}
where $\lambda_1 = 0.0$ and $\lambda_2 = 2.2$ for $f_1$ and $f_2$. All parameters are limited to the range $[-2, 2]$, with the feature space defined by the first two parameters.

\textit{ZDT3}.
When spread across the objective space is desired, objectives themselves could be used as features. This use case is demonstrated with the ZDT3 benchmark, a 30 variable problem from the ZDT MOO benchmark problem suite ~\cite{zitzler2000comparison} whose hallmark is a set of disconnected Pareto-optimal fronts, and whose first parameter is value of the first objective. Parameter ranges span 0-1 with the first two parameters used as features, enforcing a spread of solutions across the range of the first objective.

\textit{DTLZ3}. To illustrate T-DominO's bias toward balanced solutions we analyze its performance on DTLZ3, a many-objective benchmark with a tunable number of objectives and variables\cite{deb2002scalable}. We test with 10 parameters and 5 objectives, with the 6th and 7th parameters use as features.\footnote{The first $n$ parameters are explicitly linked to the first $n$ objectives as in ZDT3 -- later parameters are used to avoid explicitly exploring the objective space.}.

\subsubsection{Baseline Approaches}\hfill\\
\indent\textit{ME Single.} MAP-Elites~\cite{mapelites} optimizing only a single objective is used to establish an upper and lower bound of performance we can expect from MAP-Elites. Blind to the second objective we can expect it to find the top performing solutions for the first. Equally important, the exploration of all bins without regard to the performance on the second objective establishes a floor for performance -- the performance we could expect for having any solution in the bin.


\textit{ME Sum.} 
We compare using the T-Domino objective with MAP-Elites~\cite{mapelites} using the most naive way of combining multiple objective -- simply adding them. Our benchmarks all have well-scaled objectives, but this is typically not the case. To simulate this difficulty we use a weighted sum, with each additional objective values increased by an order of magnitude (e.g $\times$1, $\times$10, $\times$100...). 

\textit{NSGA-II.}
NSGA-II~\cite{nsga2} is used as a benchmark for conventional multi-objective optimization without feature space exploration, reaching near the Pareto front on these simple benchmarks. Though it is not our goal to compete with MOO algorithms, they provide a useful metric to contextualize the difference between exploratory approaches and pure optimizers.

\subsubsection{Settings.} In all MAP-Elites approaches the feature space is partitioned a 20x20 grid, with 2 CMA-ME improvement emitters~\cite{cmame} performing optimization. T-Domino was computed using the neighbors from 4 bins away, using a history of the 10 most recent elites in each bin. Hyperparameters for NSGA-II were kept comparable, a population of 400 matched the 400 bins of the MAP-Elites grids, with the same number of new solutions generated per generation for the same number of generations. A standard implementation of NSGA-II from the PyMoo library~\cite{pymoo} is used, as well as the library's formulations for the ZDT3 and DTLZ benchmarks whose the exact formulation is included in the Supplemental \ref{sssec:zdt3}. The PyRibs~\cite{pyribs} library was used as a basis for all MAP-Elites experiments, with T-DominO implemented as a specialized archive type. All experiments were replicated 30 times, additional plots are provided in the Supplemental.

    \subsection{Result}        
        \begin{figure}[th!]
  \vspace{-0.2cm}
  \includegraphics[width=1.5\textwidth,center]{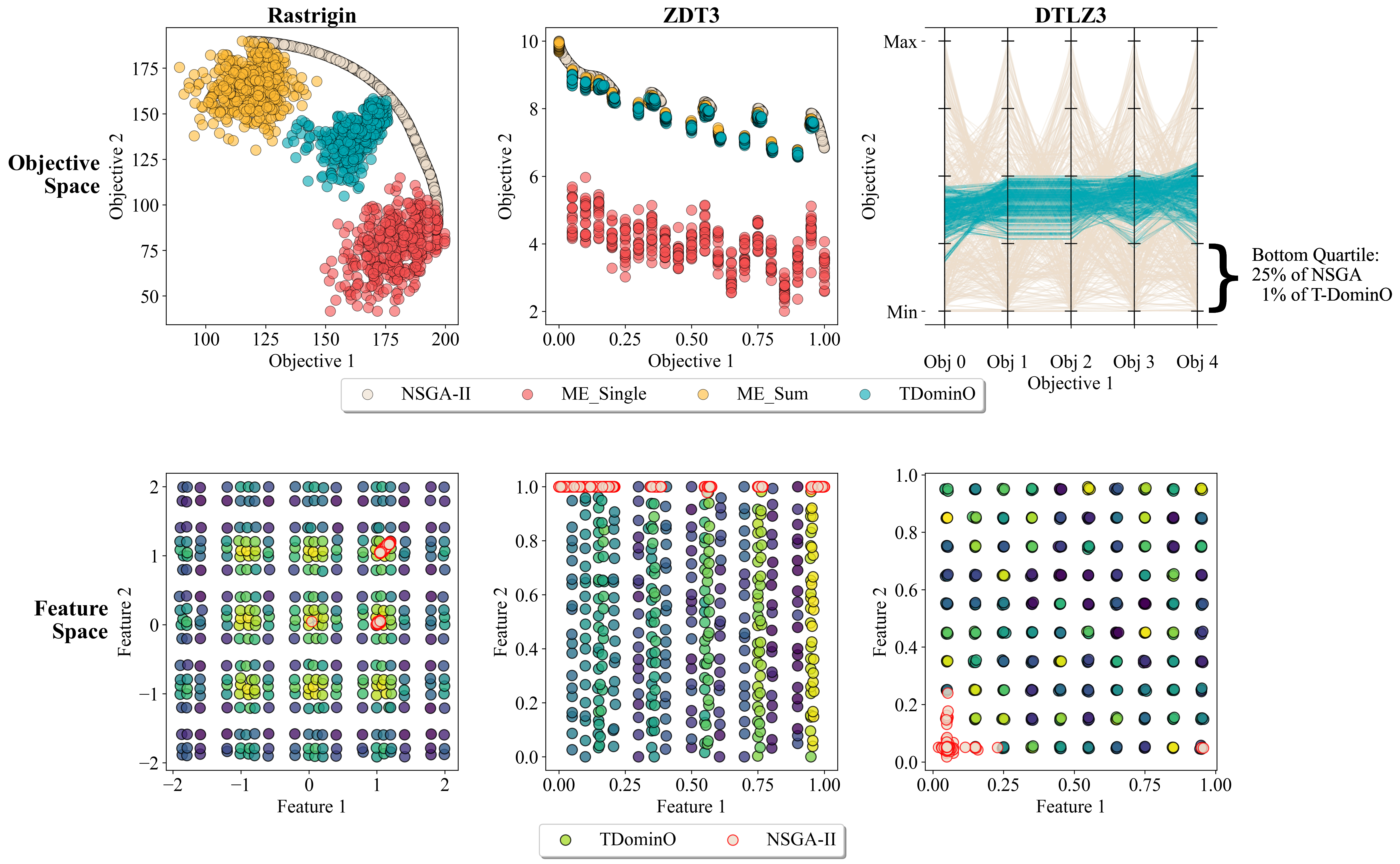}
  \caption
  { 
    \textbf{Benchmark results.} 
    \textit{Top:} Objective space as explored by each approach.
    \textit{Bottom:} Feature space as explored by T-DominO MAP-Elites and NSGA-II. T-DominO points are colored by T-DominO score compute with the entire archive as anchor points.
  }
  \label{fig:benchmark}
\end{figure}



Figure \ref{fig:benchmark} illustrates the explored regions of objective and feature space in a single run. Using the NSGA-II solutions to outline the true Pareto front we can see where each MAP-Elites approach concentrates. In the RastriginMOO case, though each version of MAP-Elites explores identical areas of the feature space, the range of possible values in objective space is large. T-Domino produces solutions in the middle of the front, with solutions that strike a balance between the two objectives. In the ZDT3 case we see that by explicitly exploring one of the objectives we can force spread over the objective space, and provide high performing solutions in the other objective.

With more than two objectives the balance seeking property of T-DominO becomes even more pronounced. Parallel coordinate plots (Figure \ref{fig:benchmark}, top right), which plot each solution as a line with one vertex per objective, make clear the differing selection pressure of T-DominO and non-dominated sorting. In contrast to the spiky lines denoting high performance on some objectives and low performance on others, T-DominO's solutions form a flat band of even performance. 

The difference of balance is critical when approaching MCX problems. To spread across the five dimensional front, solutions found by NSGA-II must span many areas with solutions that perform poorly on some objectives. If we divide the range of objective values found by NSGA-II into quartiles, only 25\% of the solutions found by NSGA-II perform over the bottom quartile on all objectives. If all of these objectives are valued by the user, that means that three quarters of solutions may be discarded immediately -- and this will only worsen as the number of objectives grows. In contrast, when T-DominO MAP-Elites' results are judged on the same scale, 99\% of the solutions found by T-DominO MAP-Elites perform over the bottom quartile on all objectives.

Visualizing the distribution of found solutions in feature space (Figure \ref{fig:benchmark}, bottom) gives a stark illustration of the main motivation for using a QD approach. The solutions produced by NSGA-II cluster in a tiny portion of the feature space. This region may be Pareto optimal, but QD gives us the ability to explore areas of our choosing.

\section{Case Study}
        \subsection{Setup}
        \begin{figure}[t]
  \centering
  \vspace{-0.2cm}
  \includegraphics[width=\figscale\textwidth,center]{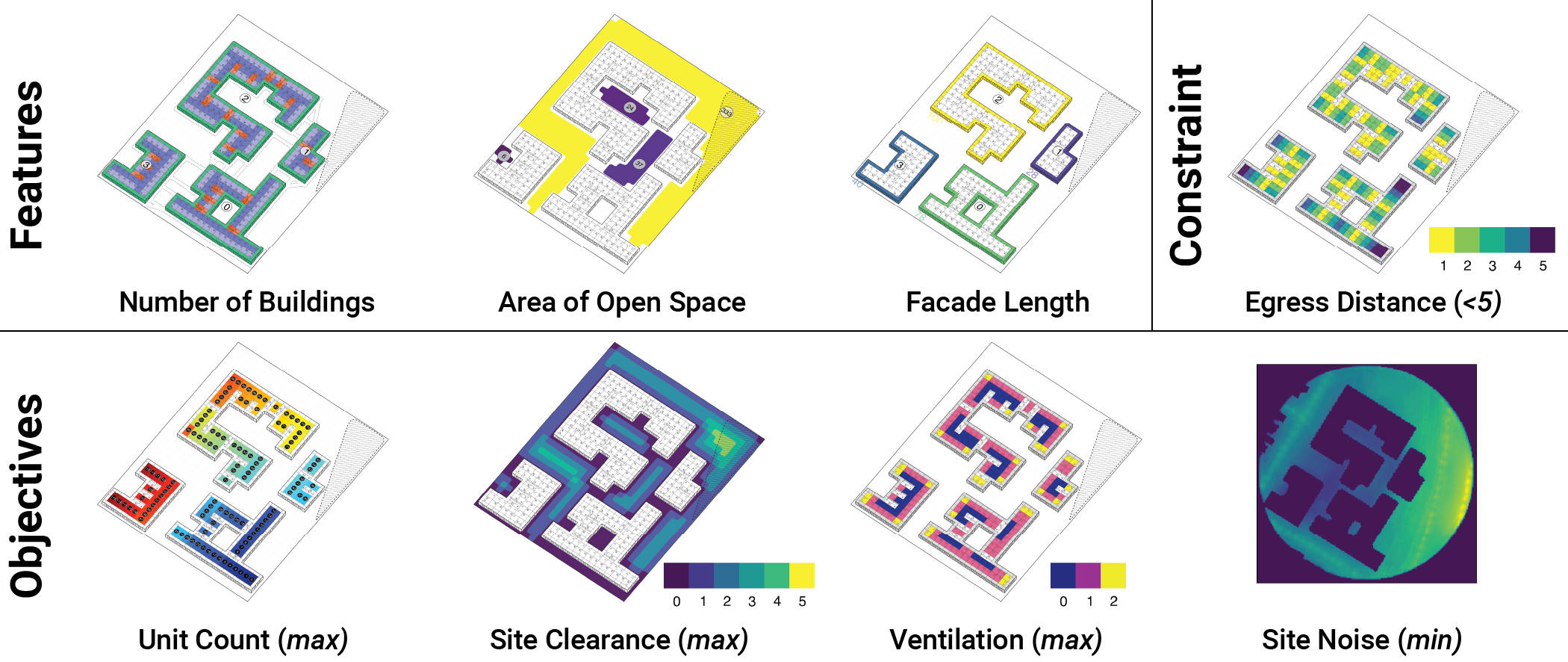}
  \caption
  { 
    \textbf{Objectives, Features, and Constraints of building layout study.} Shaded regions indicate portion of building site which cannot be built on. For details on computation of each metric see Table \ref{tab:wfc_obj} in the Supplemental. 
  }
  \label{fig:wfc_obj}
\end{figure}
\begin{figure}[th!]
  \centering
  \vspace{-0.2cm}
  \includegraphics[width=\figscale\textwidth,center]{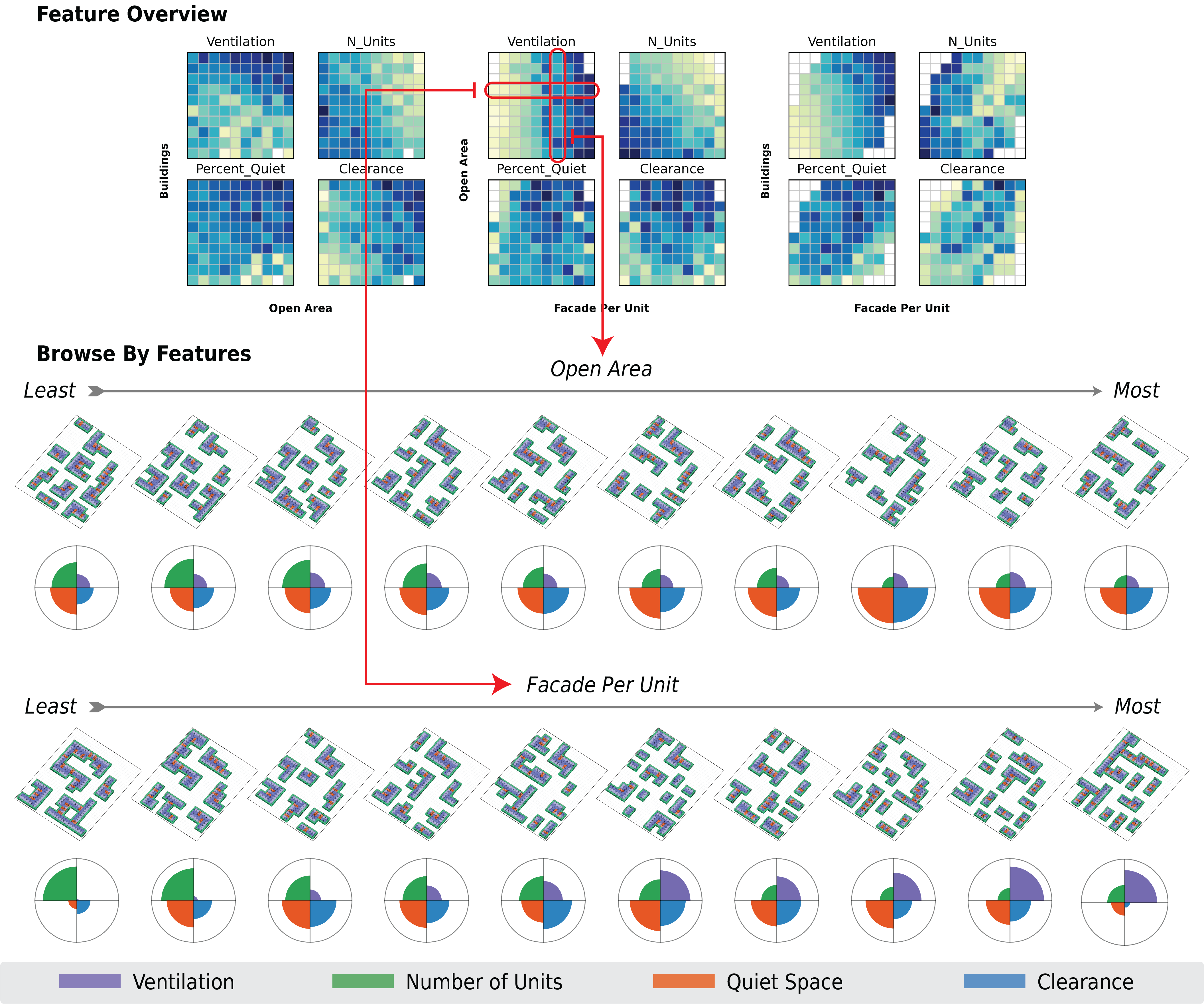}
  \caption
  { 
    \textbf{Exploring building layouts generated with T-DominO MAP-Elites.} Top: 2D views of 3D feature space, solutions in groups of four are identical, colored in each view by each objective (darker is better). Bottom: A walk through designs that vary along one feature dimension, with accompanying objective values. Petal plots are scaled to the final min/max objective value found in the archive.
  }
  \label{fig:wfc_designs}
\end{figure}


As a study of the applicability of T-DominO MAP-Elites to MCX problems we explore its use in optimizing building layouts for real-world residential complex. Solutions are produced using wave function collapse (WFC)~\cite{wfc}, a popular tool for tile-based procedural content generation in games. WFC is a constraint satisfaction approach which extracts local patterns from a small set of samples, and transforms them into a set of local constraints. The constraints drive generation, ensuring that every local patch of the output also exists in the set of input examples. We adapt the implementation here~\cite{karth2019addressing,wfccode}. Constrained generation systems like WFC are particularly appropriate for the semi-constrained design systems often used in residential building, such as modular or prefabricated units, and do not require the extensive curated datasets of valid designs.

WFC, though constrained, is a stochastic process. At every iteration a tile is `collapsed', or fixed, with the type chosen stochastically from a list of valid tiles, and new constraints applied to its neighbors. To make this encoding more amenable to optimization we have introduced an evolvable genotype of tiles which are fixed at the beginning of this collapsing process. Children inherit these fixed tiles from parents, in addition to fixing an additional tile from the design produced by the parent or removing one of the tiles that were fixed by the parent. Fixing tiles freezes key portions of the parent design and saves progress toward interesting designs – while still allowing substantial deviation from the parent, as the remainder of the tiles are generated stochastically with WFC. See the supplemental material for set of used tiles and example designs.

The resulting designs are evaluated according to 4 objectives, 1 constraint, and explored along 3 features (illustrated in Figure \ref{fig:wfc_obj}, more details in Table \ref{tab:wfc_obj} in the Supplemental). The constraint was handled in a tournament fashion as in \cite{deb2000efficient} -- in any tournament where one solution follows the constraint and the other does not, the solution which follows the constraint wins regardless.

        \subsection{Result}
        
Once a set of designs has been produced, we can extend on MAP-Elites' intuitive way of optimizing, organizing, and displaying solutions to multiple objectives. 
To better understand our 3D feature space we `flatten' it into a set of 2D views
by creating a set of new 2D archives with the desired feature axes and inserting all of the solutions from the 3D archive, forcing competition based on only two features. The result is one map for each pair of the three features (Figure \ref{fig:wfc_designs}, top).

Each of these views can in turn be split into one map for each objective, collectively allowing correlations along feature axes and objectives to all be seen at a glance. Obvious relationships such as an increase in open area resulting in more units are clear, along with less appreciated connections: with fewer buildings ventilation is worse -- unless buildings have longer facades.

Clear organized grids of solutions open up many avenues for intuitive navigation of the produced solutions. Here we show one possibility, browsing rows or columns of designs. An area of the map can be selected, and the individual designs displayed along with their performance across objectives. Drilling down on a subset we can see the qualitative differences between large and small open areas (Figure \ref{fig:wfc_designs}, Center), and the kinds of layouts they each represent. A combined view allows us to see large difference in objective values that may not have been apparent from a qualitative glance: even with the same amount of open area we can see there is huge amount of variation in the number of units that can fit on a site, that increasing this amount of units comes at the cost of natural ventilation, and that qualitatively this trade-off is between a few thick buildings, or several small ones (Figure \ref{fig:wfc_designs}, Bottom).

\section{Discussion}


In this work we have defined a new class of problem, the MCX problem, tailored specifically to the needs of generative design. A chief aim of the generative design is to spark ideas and explore concepts, and results are typically explored by browsing designs not objectives. The measure space of MAP-Elites provides an intuitive way of creating and exploring sets of solutions with varied and understandable high level features.

T-DominO allows MAP-Elites to maintain these visual and organizational capabilities in the complex multi-criteria scenarios where they are most useful. Keeping a single solution in each bin rather than a front is about more than computational cost, it is about maintaining visual accessibility. Having a single solution in each bin simplifies browsing and selecting interesting designs. When objectives and features are correlated, the possible objectives values for each feature combination is constrained to a range -- so though balanced solutions are found, the larger pattern of objective/feature relations are still clear.

T-DominO allows us to optimize for multiple objectives in a QD setting without making any other fundamental changes to the algorithm. Simple to implement, without adding any appreciable computational burden, T-DominO can be easily integrated into existing approaches. By leaving its elitist character untouched, T-Domino allows MAP-Elites to handle multiple objectives while maintaining its core visualization and presentation strengths. Equipping MAP-Elites with T-DominO allows us to generate diverse sets of well-rounded high performing solutions, creating a powerful tool for tackling MCX problems. 


\section{Acknowledgements}
        The authors would like to thank Renaud Danhaive, Jeffrey Landes, and the entire Spacemaker team for their invaluable site analysis tool and expertise as well as Mark Davis and David Benjamin for their guidance and support.         
        
\section{Supplemental Material and Code}
Supplemental material and code available at:\\ \href{https://github.com/agaier/tdomino_ppsn}{https://github.com/agaier/tdomino\_ppsn}

%
\bibliographystyle{splncs04}
\bibliography{main.bib} 

\begin{thebibliography}{10}
\providecommand{\url}[1]{\texttt{#1}}
\providecommand{\urlprefix}{URL }
\providecommand{\doi}[1]{https://doi.org/#1}

\bibitem{alvarez2019empowering}
Alvarez, A., Dahlskog, S., Font, J., Togelius, J.: Empowering quality diversity
  in dungeon design with interactive constrained map-elites. In: 2019 IEEE
  Conference on Games (CoG). pp.~1--8. IEEE (2019)

\bibitem{arieff2013new}
Arieff, A.: New forms that function better. Technology Review pp. 94--98 (2013)

\bibitem{pymoo}
{Blank}, J., {Deb}, K.: pymoo: Multi-objective optimization in python. IEEE
  Access  \textbf{8},  89497--89509 (2020)

\bibitem{bradner2014parameters}
Bradner, E., Iorio, F., Davis, M.: Parameters tell the design story: ideation
  and abstraction in design optimization. In: Proceedings of the symposium on
  simulation for architecture \& urban design. vol.~26. Society for Computer
  Simulation International (2014)

\bibitem{charity2020mech}
Charity, M., Green, M.C., Khalifa, A., Togelius, J.: Mech-elites: Illuminating
  the mechanic space of gvg-ai. In: International Conference on the Foundations
  of Digital Games. pp. 1--10 (2020)

\bibitem{cully2021multi}
Cully, A.: Multi-emitter map-elites: improving quality, diversity and data
  efficiency with heterogeneous sets of emitters. In: Proceedings of the
  Genetic and Evolutionary Computation Conference. pp. 84--92 (2021)

\bibitem{cully2015robots}
Cully, A., Clune, J., Tarapore, D., Mouret, J.B.: Robots that can adapt like
  animals. Nature  \textbf{521}(7553),  503--507 (2015)

\bibitem{cully2017quality}
Cully, A., Demiris, Y.: Quality and diversity optimization: A unifying modular
  framework. IEEE Transactions on Evolutionary Computation  \textbf{22}(2),
  245--259 (2017)

\bibitem{deb2000efficient}
Deb, K.: An efficient constraint handling method for genetic algorithms.
  Computer methods in applied mechanics and engineering  \textbf{186}(2-4),
  311--338 (2000)

\bibitem{moo}
Deb, K.: Multi-objective optimization. In: Search methodologies, pp. 403--449.
  Springer (2014)

\bibitem{nsga3}
Deb, K., Jain, H.: An evolutionary many-objective optimization algorithm using
  reference-point-based nondominated sorting approach, part i: solving problems
  with box constraints. IEEE transactions on evolutionary computation
  \textbf{18}(4),  577--601 (2013)

\bibitem{nsga2}
Deb, K., Pratap, A., Agarwal, S., Meyarivan, T.: A fast and elitist
  multiobjective genetic algorithm: Nsga-ii. IEEE transactions on evolutionary
  computation  \textbf{6}(2),  182--197 (2002)

\bibitem{deb2002scalable}
Deb, K., Thiele, L., Laumanns, M., Zitzler, E.: Scalable multi-objective
  optimization test problems. In: Proceedings of the 2002 Congress on
  Evolutionary Computation. CEC'02 (Cat. No. 02TH8600). vol.~1, pp. 825--830.
  IEEE (2002)

\bibitem{cmame}
Fontaine, M.C., Togelius, J., Nikolaidis, S., Hoover, A.K.: Covariance matrix
  adaptation for the rapid illumination of behavior space. In: Proceedings of
  the 2020 genetic and evolutionary computation conference. pp. 94--102 (2020)

\bibitem{sail_aiaa}
Gaier, A., Asteroth, A., Mouret, J.B.: Aerodynamic design exploration through
  surrogate-assisted illumination. In: 18th AIAA/ISSMO multidisciplinary
  analysis and optimization conference. p.~3330 (2017)

\bibitem{sail_gecco}
Gaier, A., Asteroth, A., Mouret, J.B.: Data-efficient exploration,
  optimization, and modeling of diverse designs through surrogate-assisted
  illumination. In: Proceedings of the Genetic and Evolutionary Computation
  Conference. pp. 99--106 (2017)

\bibitem{sail_ecj}
Gaier, A., Asteroth, A., Mouret, J.B.: Data-efficient design exploration
  through surrogate-assisted illumination. Evolutionary computation
  \textbf{26}(3),  381--410 (2018)

\bibitem{gerber2012design}
Gerber, D.J., Lin, S.H., Pan, B., Solmaz, A.S.: Design optioneering:
  multi-disciplinary design optimization through parameterization, domain
  integration and automation of a genetic algorithm. In: Proceedings of the
  2012 Symposium on Simulation for Architecture and Urban Design. pp.~1--8
  (2012)

\bibitem{gonzalez2020finding}
Gonz{\'a}lez-Duque, M., Palm, R.B., Ha, D., Risi, S.: Finding game levels with
  the right difficulty in a few trials through intelligent trial-and-error. In:
  2020 IEEE Conference on Games (CoG). pp. 503--510. IEEE (2020)

\bibitem{gravina2019procedural}
Gravina, D., Khalifa, A., Liapis, A., Togelius, J., Yannakakis, G.N.:
  Procedural content generation through quality diversity. In: 2019 IEEE
  Conference on Games (CoG). pp.~1--8. IEEE (2019)

\bibitem{greiner2007improving}
Greiner, D., Emperador, J.M., Winter, G., Galv{\'a}n, B.: Improving
  computational mechanics optimum design using helper objectives: an
  application in frame bar structures. In: International Conference on
  Evolutionary Multi-Criterion Optimization. pp. 575--589. Springer (2007)

\bibitem{wfc}
Gumin, M.: Bitmap and tilemap generation from a single example by collapsing a
  wave function. GitHub (2016)

\bibitem{produqd}
Hagg, A., Asteroth, A., B{\"a}ck, T.: Prototype discovery using
  quality-diversity. In: International Conference on Parallel Problem Solving
  from Nature. pp. 500--511. Springer (2018)

\bibitem{sphen}
Hagg, A., Wilde, D., Asteroth, A., B{\"a}ck, T.: Designing air flow with
  surrogate-assisted phenotypic niching. In: International Conference on
  Parallel Problem Solving from Nature. pp. 140--153. Springer (2020)

\bibitem{handl2008multiobjectivization}
Handl, J., Lovell, S.C., Knowles, J.: Multiobjectivization by decomposition of
  scalar cost functions. In: International Conference on Parallel Problem
  Solving from Nature. pp. 31--40. Springer (2008)

\bibitem{holzer2007parametric}
Holzer, D., Hough, R., Burry, M.: Parametric design and structural optimisation
  for early design exploration. International Journal of Architectural
  Computing  \textbf{5}(4),  625--643 (2007)

\bibitem{jensen2004helper}
Jensen, M.T.: Helper-objectives: Using multi-objective evolutionary algorithms
  for single-objective optimisation. Journal of Mathematical Modelling and
  Algorithms  \textbf{3}(4),  323--347 (2004)

\bibitem{wfccode}
Karth, I.: wfc2019f. \url{https://github.com/ikarth/wfc-2019f} (2021)

\bibitem{karth2019addressing}
Karth, I., Smith, A.M.: Addressing the fundamental tension of pcgml with
  discriminative learning. In: Proceedings of the 14th International Conference
  on the Foundations of Digital Games. pp.~1--9 (2019)

\bibitem{knowles2001reducing}
Knowles, J.D., Watson, R.A., Corne, D.W.: Reducing local optima in
  single-objective problems by multi-objectivization. In: International
  conference on evolutionary multi-criterion optimization. pp. 269--283.
  Springer (2001)

\bibitem{matejka2018dream}
Matejka, J., Glueck, M., Bradner, E., Hashemi, A., Grossman, T., Fitzmaurice,
  G.: Dream lens: Exploration and visualization of large-scale generative
  design datasets. In: Proceedings of the 2018 CHI Conference on Human Factors
  in Computing Systems. pp. 1--12 (2018)

\bibitem{mouret2011novelty}
Mouret, J.B.: Novelty-based multiobjectivization. In: New horizons in
  evolutionary robotics, pp. 139--154. Springer (2011)

\bibitem{mapelites}
Mouret, J.B., Clune, J.: Illuminating search spaces by mapping elites. arXiv
  preprint arXiv:1504.04909  (2015)

\bibitem{nagy2018generative}
Nagy, D., Villaggi, L., Benjamin, D.: Generative urban design: Integration of
  financial and energy design goals in a generative design workflow for
  residential neighborhood layout. In: Symposium on Simulation for Architecture
  and Urban Design (2018)

\bibitem{nagy2017project}
Nagy, D., Lau, D., Locke, J., Stoddart, J., Villaggi, L., Wang, R., Zhao, D.,
  Benjamin, D.: Project discover: An application of generative design for
  architectural space planning. In: Proceedings of the Symposium on Simulation
  for Architecture and Urban Design. p.~7. Society for Computer Simulation
  International (2017)

\bibitem{nagy2017beyond}
Nagy, D., Villaggi, L., Zhao, D., Benjamin, D.: Beyond heuristics: a novel
  design space model for generative space planning in architecture  (2017)

\bibitem{panichella2019adaptive}
Panichella, A.: An adaptive evolutionary algorithm based on non-euclidean
  geometry for many-objective optimization. In: Proceedings of the Genetic and
  Evolutionary Computation Conference. pp. 595--603 (2019)

\bibitem{moo_qd}
Pierrot, T., Richard, G., Beguir, K., Cully, A.: Multi-objective quality
  diversity optimization. arXiv preprint arXiv:2202.03057  (2022)

\bibitem{pugh2016quality}
Pugh, J.K., Soros, L.B., Stanley, K.O.: Quality diversity: A new frontier for
  evolutionary computation. Frontiers in Robotics and AI  \textbf{3}, ~40
  (2016)

\bibitem{pyribs}
Tjanaka, B., Fontaine, M.C., Zhang, Y., Sommerer, S., Dennler, N., Nikolaidis,
  S.: pyribs: A bare-bones python library for quality diversity optimization.
  \url{https://github.com/icaros-usc/pyribs} (2021)

\bibitem{turrin2011design}
Turrin, M., Von~Buelow, P., Stouffs, R.: Design explorations of performance
  driven geometry in architectural design using parametric modeling and genetic
  algorithms. Advanced Engineering Informatics  \textbf{25}(4),  656--675
  (2011)

\bibitem{zitzler2000comparison}
Zitzler, E., Deb, K., Thiele, L.: Comparison of multiobjective evolutionary
  algorithms: Empirical results. Evolutionary computation  \textbf{8}(2),
  173--195 (2000)

\end{thebibliography}

\newpage
\section{Supplemental Material}
Upon publication all supplemental material, along with all source code used to produce the results in this paper will be published online.

\subsection{Building Layout Objectives, Features, and Constraints}
\begin{table}[h]
\begin{tabular}{p{0.20\linewidth}  p{0.80\linewidth}}
\hline
\textbf{Objectives}    & \textbf{Description} \\

\textit{Ventilation}   &  Natural indoor ventilation potential for each apartment. Computed as a simplified version of the air flow network (AFN) methodology: computing the connectivity distance of each room to the apartment’s windows.                   \\

\textit{Clearance}     & The landscaping capacity for carbon sequestration: measuring the potential capacity for outdoor green areas to store and avoid carbon [i-Tree, 2021]. For this case study we included a simplified version that limits its computation solely to a clearance metric, which is the amount of clear space that green areas have from adjacent buildings to promote the growth of carbon storing trees.                     \\

\textit{Site Noise}    & Defined as the percentage of tiles on the site with a noise level of less than 50db, as estimated by analysis of the real-world site and a surrogate model trained on a large set of noise analysis simulations performed on apartment complex designs designed manually by customers.                     \\

\textit{Units}         & The number of apartment units in the complex.                     \\ \hline

\textbf{Measures}      &                      \\

\textit{Facade Length} & Ratio of number of facade (border) tiles to number of units.                     \\

\textit{Buildings}     & Count of buildings.                     \\

\textit{Open Area}     & Surface area of open spaces. Open spaces are
computed as the number of empty tiles not occupied by buildings after performing an erosion operation to remove narrow corridors between buildings.
                     \\ \hline

\textbf{Constraints}   &                      \\

\textit{Egress}        & Tile distance of each apartment unit to the building’s cores, which contain stairs. The distance to a core must be less than 5 tiles.                    
\end{tabular}\label{tab:wfc_obj}
\end{table}

\newpage
\subsection{Wave Function Collapse Tiles Set and Seed Examples}
\begin{figure}[h!]
  \centering
  \vspace{-0.2cm}
  \includegraphics[width=\figscale\textwidth,center]{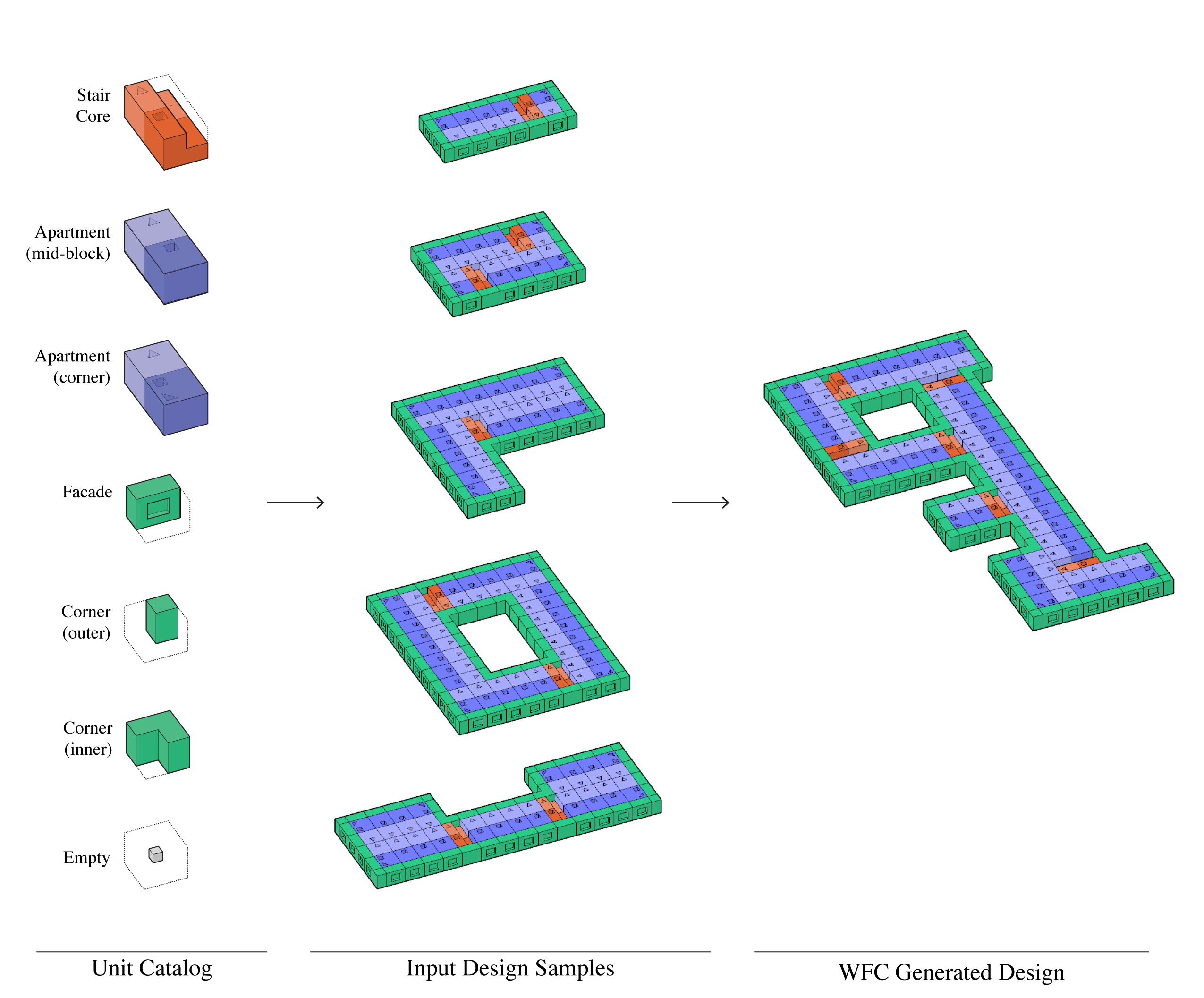}
  \caption
  { 
    \textbf{Tiles set and example designs used to seed Wave Function Collapse.}
  }
  \label{fig:tiles}
\end{figure}

\newpage
\subsection{Single Building Example Outputs of Wave Function Collapse}
\begin{figure}[h!]
  \centering
  \vspace{-0.2cm}
  \includegraphics[width=\figscale\textwidth,center]{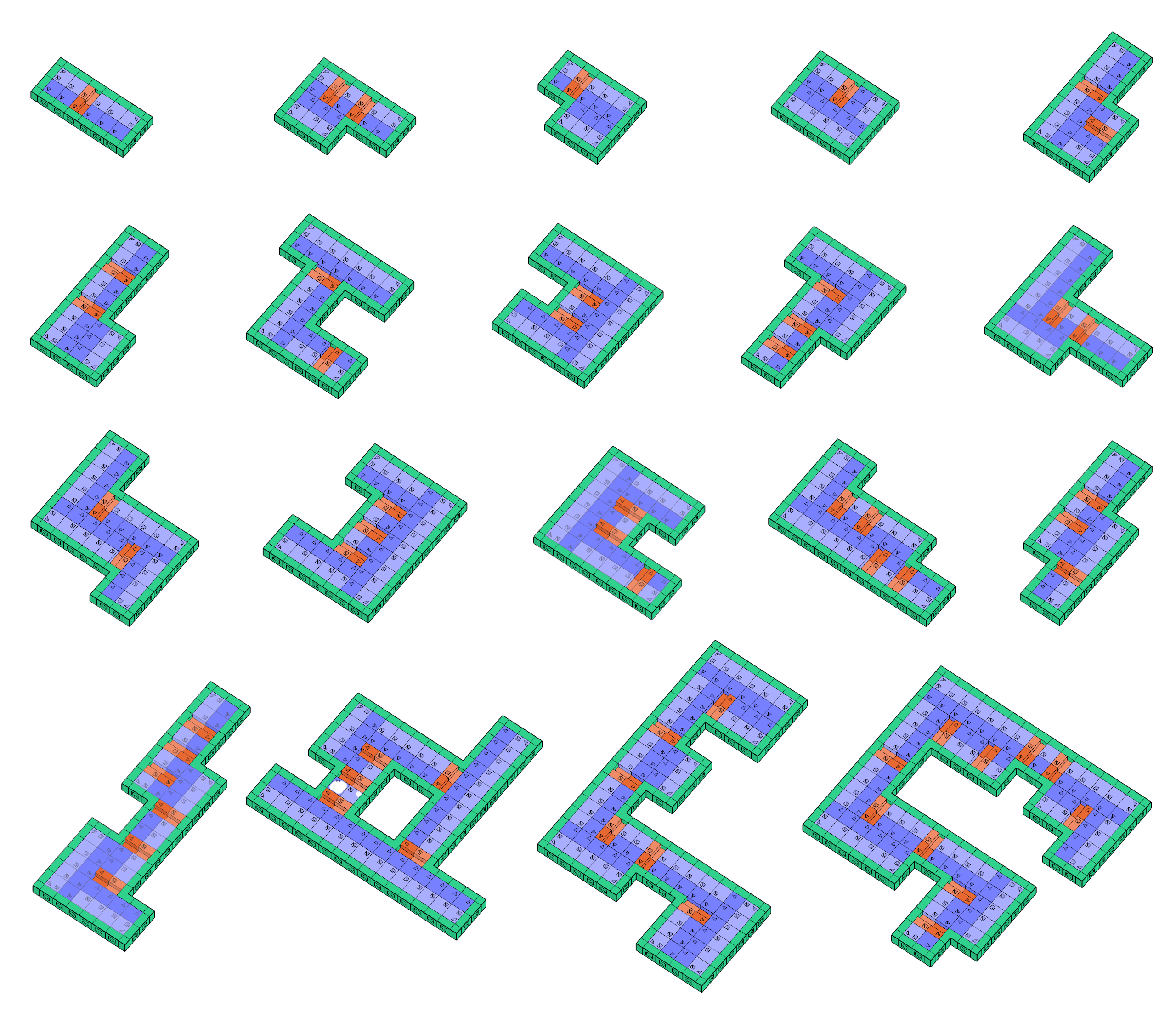}
  \caption
  { 
    \textbf{Example Buildings Produced Using Wave Function Collapse.}
  }
  \label{fig:examples}
\end{figure}

\newpage
\subsection{QD Score}
\begin{figure}[h!]
  \centering
  \vspace{-0.2cm}
  \includegraphics[width=\figscale\textwidth,center]{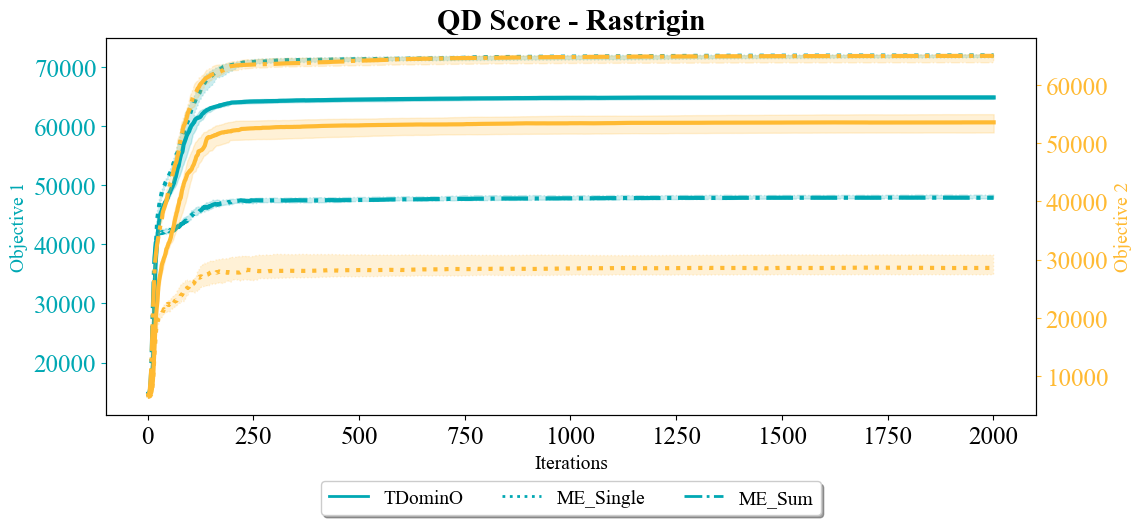}
  \includegraphics[width=\figscale\textwidth,center]{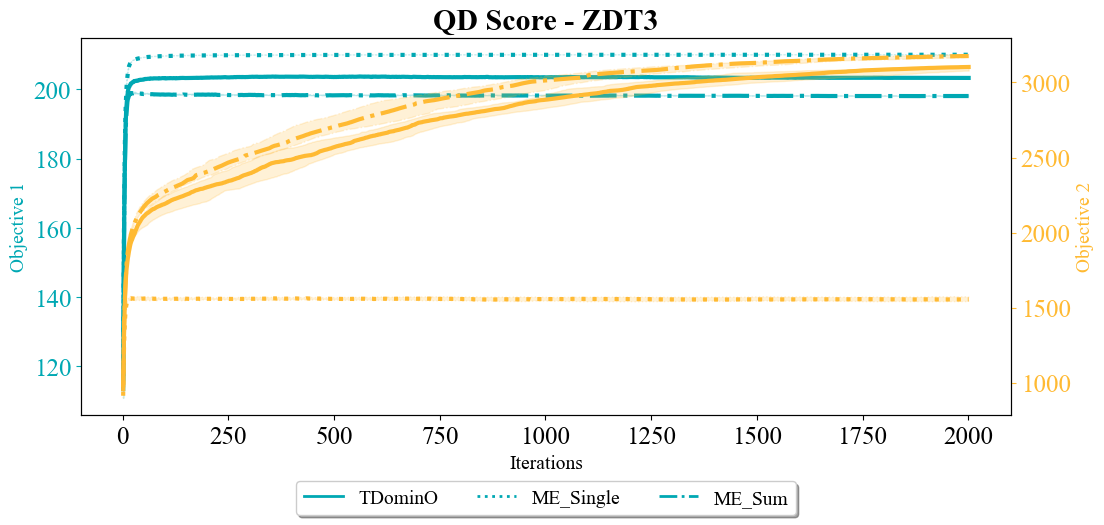}
  \caption
  { 
    \textbf{QD Scores on benchmarks.} Shaded regions indicated border of top and bottom quartile over 30 runs.
  }
  \label{fig:examples}
\end{figure}

\newpage
\subsection{MOO Benchmark Functions}

The following benchmark functions were used in this work. For further details and the exact Python implementations used see: \\ \url{https://pymoo.org/problems/test\_problems.html}

\subsubsection{ZDT3}\label{sssec:zdt3}
The ZDT3 benchmark objective function is defined as:

$
\begin{aligned}
f_{1}(x) &=x_{1} \\
g(x) &=1+\frac{9}{n-1} \sum_{i=2}^{n} x_{i} \\
h\left(f_{1}, g\right) &=1-\sqrt{f_{1} / g}-\left(f_{1} / g\right) \sin \left(10 \pi f_{1}\right) \\
0 & \leq x_{i} \leq 1 \quad i=1, \ldots, n
\end{aligned}
$

\subsubsection{DTLZ3}\label{sssec:dltz3}
The DTLZ3 benchmark objective function is defined as:

Min. $f_{1}(\mathbf{x})=\left(1+g\left(\mathbf{x}_{M}\right)\right) \cos \left(x_{1} \pi / 2\right) \cdots \cos \left(x_{M-2} \pi / 2\right) \cos \left(x_{M-1} \pi / 2\right)$,

Min. $f_{2}(\mathbf{x})=\left(1+g\left(\mathbf{x}_{M}\right)\right) \cos \left(x_{1} \pi / 2\right) \cdots \cos \left(x_{M-2} \pi / 2\right) \sin \left(x_{M-1} \pi / 2\right)$,

Min. $f_{3}(\mathbf{x})=\left(1+g\left(\mathbf{x}_{M}\right)\right) \cos \left(x_{1} \pi / 2\right) \cdots \sin \left(x_{M-2} \pi / 2\right)$,

$\vdots \quad \vdots$

Min. $f_{M}(\mathbf{x})=\left(1+g\left(\mathbf{x}_{M}\right)\right) \sin \left(x_{1} \pi / 2\right)$,

with $g\left(\mathbf{x}_{M}\right)=100\left[\left|\mathbf{x}_{M}\right|+\sum_{x_{i} \in \mathbf{x}_{M}}\left(x_{i}-0.5\right)^{2}-\cos \left(20 \pi\left(x_{i}-0.5\right)\right)\right]$,
$0 \leq x_{i} \leq 1, \quad$ for $i=1,2, \ldots, n$.

\end{document}